\documentclass[11pt]{article}
\usepackage{eamt26}
\usepackage{times}
\usepackage{url}
\usepackage{multirow}
\usepackage{latexsym}
\usepackage[small,bf]{caption}
\usepackage{pgfplots}
\usepackage{pgffor}
\usepgfplotslibrary{external}
\pgfdeclarelayer{background}
\pgfsetlayers{background,main}
\usepackage{booktabs}
\usepackage{graphicx}
\usepackage{listings}
\usepackage{tabularx}
\usepackage{array}

\lstdefinestyle{promptstyle}{
  basicstyle=\ttfamily\footnotesize,
  columns=fullflexible,
  breaklines=true,
  frame=single,
  framesep=0.6em,
  framerule=0.4pt,
  xleftmargin=0.8em,
  xrightmargin=0.8em,
  showstringspaces=false,
  keepspaces=true,
  lineskip=0.2em,
  aboveskip=1em,
  belowskip=1em
}

\title{Translating Under Pressure: Domain-Aware LLMs for Crisis Communication}

\author{Antonio Castaldo\\
  University of Pisa\\
  University of Naples ``L'Orientale''\\
  {\tt antonio.castaldo@phd.unipi.it}  \And
  Maria Carmen Staiano\\
  University of Macerata\\
  {\tt m.staiano@unimc.it}
  \AND
  Johanna Monti\\
  University of Naples ``L'Orientale''\\
  {\tt jmonti@unior.it} \And
  Sheila Castilho\\
  Dublin City University\\
  {\tt sheila.castilho@dcu.ie}
  \AND
  Francesca Chiusaroli\\
  University of Macerata\\
  {\tt f.chiusaroli@unimc.it}
}

\date{}

\begin{document}
\maketitle

\begin{abstract}
Timely and reliable multilingual communication is critical during natural and human-induced disasters, but developing effective solutions for crisis communication is limited by the scarcity of curated parallel data. We propose a domain-adaptive pipeline that expands a small reference corpus, by retrieving and filtering data from general corpora. We use the resulting dataset to fine-tune a small language model for crisis-domain translation and then apply preference optimization to bias outputs toward CEFR A2-level English. Automatic and human evaluation shows that this approach improves readability, while maintaining strong adequacy. Our results indicate that simplified English, combined with domain adaptation, can function as a practical lingua franca for emergency communication when full multilingual coverage is not feasible. 
\end{abstract}

\section{Introduction}

Global crises such as the COVID-19 pandemic, wildfires, or earthquakes require the rapid, trustworthy, and reliable dissemination of public information \cite{piller2020linguistic,hajek2024understanding}. In linguistically diverse contexts such as Italy, where international residents and visitors may have limited proficiency in Italian, ensuring access to emergency communications becomes a matter of public safety. When full multilingual coverage across all community languages is not feasible, simplified English can serve as a practical lingua franca, enabling broader comprehension and supporting timely action among individuals with different linguistic backgrounds \cite{musacchio2017localising,o2020crisis,radicioni2022cultural}.

In this context, machine translation (MT) systems play a critical role in assisting emergency response teams and public authorities in disseminating information efficiently \cite{lewis2010haitian}. The emergence of Large Language Models (LLMs) has further expanded the potential of MT in crisis settings \cite{lankford2024leveraging}. Their adaptability to specific domains and communicative requirements makes them promising tools for producing emergency messages tailored to diverse audiences.

However, effective risk communication is not solely a matter of semantic accuracy. Emergency messages must be actionable, terminologically precise, and easily understandable, particularly for non-native readers. This requires careful handling of domain-specific terminology alongside explicit attention to readability. Achieving this balance depends on carefully curated parallel corpora that support reliable domain adaptation \cite{moorkens2025sociotechnical}.

Within the Italian context, existing Institution-to-User communication resources \cite{torresi2020translating} provide valuable resources, but remain limited in size. In this study, we adopt ITALERT as our reference corpus \cite{staiano2025italert}, and use it to retrieve relevant parallel data from general corpora. ITALERT encompasses major natural disasters in Italy, and includes high-quality parallel data, making it a particularly suitable resource for our study.

In this work, we propose a domain-adaptive pipeline for generating simplified English translations of emergency messages from the Italian Civil Protection Department. We introduce a two-step methodology to retrieve and classify relevant in-domain data in the Italian-English language pair from general corpora. Using the ITALERT corpus as a reference, we employ cluster-based similarity search to construct a crisis-relevant parallel dataset. We then fine-tune a small language model on this curated corpus and further apply preference optimization to bias its outputs toward CEFR A2-level English, aiming to produce translations that are both reliable and accessible for linguistically diverse populations in Italy.

\section{Related Work}

Prior work has shown that the limited availability of appropriate in-domain datasets represents a major challenge when deploying MT in crisis settings \cite{cadwell2019more,o2022crisis,moorkens2025sociotechnical}, particularly with regard to context-sensitive translations, terminology consistency, and register accuracy. To address this problem, we elaborated a two-stage pipeline to augment crisis-domain data for a high resource language combination (Italian-English). In this context, datasets related to the crisis domain are limited, especially when taking into account information from public sources, such as government agencies. The existing parallel datasets available for our specific language combination are mostly related to public-health emergencies like COVID-19 \cite{anastasopoulos2020tico,way2020rapid}.

A strategy to mitigate data scarcity in specialized domains is to apply data retrieval techniques to mine relevant parallel segments from large general-domain corpora. Previous work has shown that semantic vector similarity, embedding-based retrieval, and classifier-driven filtering can effectively extract in-domain data for MT adaptation \cite{espana2017empirical,sugathadasa2017synergistic,glavavs2018resource}.

Research on crisis-oriented corpora has highlighted the need for structured annotation frameworks to distinguish disaster-related content from general communication \cite{alam2018crisismmd}. The taxonomy introduced by the United Nations Office for Disaster Risk Reduction and the International Science Council \shortcite{undrr_isc_hips_2025} provides an authoritative reference for hazard classification, organizing 281 hazards into eight clusters: Meteorological and Hydrological, Extraterrestrial, Geological, Environmental, Chemical, Biological, Technological, and Societal.

In this study, we use this taxonomy as a reference framework to filter in-domain sentences. Domain relevance is determined not only by explicit references to hazards but also by the presence of terminology commonly used in crisis management. Sentences may include descriptive, normative, procedural, or advisory content, and are not limited to emergency preparedness leaflets.

Research in crisis communication has also emphasized the importance of message accessibility and readability when disseminating emergency information to multilingual populations. Prior studies have shown that comprehension plays a central role in shaping public trust and behavioral compliance during crises. In particular, Rossetti et al.~\shortcite{rossetti2020comprehension} demonstrate that clearer crisis messages contribute to higher levels of trust toward emergency authorities and increase the likelihood that individuals will follow preparedness and response instructions. This line of research motivates our adoption of simplification-informed model adaptation, aimed at producing translations that are not only accurate but also comprehensible for linguistically diverse audiences.
 
Despite these advances, to the best of our knowledge, no existing work combines semantic retrieval, disaster-informed annotation, and classifier-based filtering into a unified pipeline for crisis-domain MT. 

Taken together, previous research highlights two main gaps:

(1) the lack of Italian–English crisis-domain parallel data, and
(2) the absence of combined retrieval–annotation–classification approaches for corpus augmentation.

We address these limitations by proposing a structured end-to-end pipeline that expands crisis-domain data,  fine-tunes a domain-specific MT model, and evaluates translation quality through both automatic and human-centered assessments, with the goal of improving MT reliability in crisis communication.

\section{Data Collection}

\begin{figure*}[t]
    \centering
    \includegraphics[width=\textwidth]{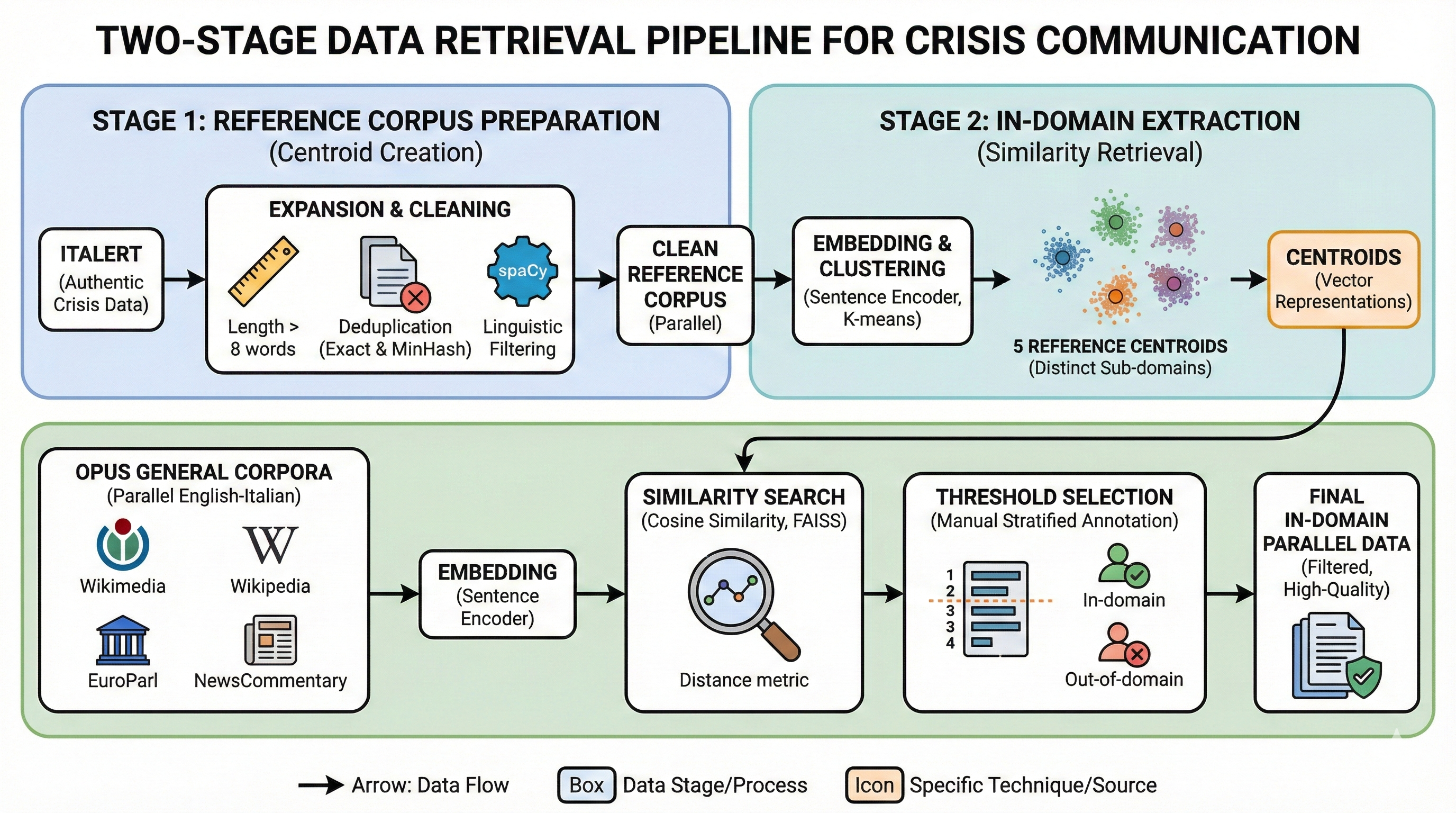} 
    \caption{Overview of our two-stage data retrieval pipeline. \textbf{Stage 1} focuses on cleaning and clustering the reference corpus to generate distinct semantic centroids. \textbf{Stage 2} leverages these centroids to retrieve in-domain sentences from general corpora (OPUS) via embedding similarity, validated by a stratified manual annotation.}
    \label{fig:pipeline}
\end{figure*}

Obtaining sufficient in-domain parallel data for risk communication remains a challenge, where communications are generally produced into monolingual corpora and rarely compiled into parallel MT resources. We address data scarcity with a two-stage approach: first assembling and expanding a small reference corpus of authentic crisis communication, starting from ITALERT \cite{staiano2025italert}, then using it to extract in-domain sentences from large general corpora, using embedding-based similarity. Our data selection pipeline is described in Figure~\ref{fig:pipeline}.

\subsection{Reference Corpus}

Our reference corpus builds upon the dataset introduced by Staiano et al.~\shortcite{staiano2025italert}. To ensure high data quality, the corpus underwent a rigorous post-processing pipeline. We applied exact and MinHash deduplication to eliminate redundancy and filtered for a minimum length ($>8$ words) to ensure sufficient context. 

This filtering was necessary to minimize semantic noise and ambiguity in the vector space. By retaining only linguistically robust examples, we ensured that the reference data formed distinct, high-density clusters, which is a prerequisite for the data selection of the following stage.

\subsection{General Corpora}

To expand our training resources beyond the limited reference corpus, we utilized the OPUS collection \cite{tiedemann_opus_2004}, specifically targeting the English-Italian language pair. We selected four sub-corpora, described in Table~\ref{tab:corpus_stats}. These sources consist of parallel sentences with human translations, and we selected them on the hypothesis that they contain high-quality crisis examples embedded within broader discourse. Although generic, these sources cover distinct facets of crisis communication, from natural disasters to legislative discussions on civil protection and disaster relief. This diversity ensures that our extraction pipeline can recover a wide spectrum of in-domain topics, from urgent alerts to procedural descriptions.

Before retrieval, the general corpora underwent rigorous preprocessing. We removed exact and fuzzy duplicates, using the MinHash deduplication algorithm. Sentences shorter than eight words were filtered out, and malformed or incomplete segments were excluded through dependency parsing with SpaCy \cite{spacy_2020}. Finally, we made sure the target corpora only contained sentences in our language pair, using langdetect.

\begin{table}[h]
\centering
\resizebox{\columnwidth}{!}{%
\begin{tabular}{lrr}
\toprule
\textbf{Source} & \textbf{Raw} & \textbf{Clean} \\
\midrule
EuroParl & 2,128,356 & 1,890,114 \\
Wikimedia & 1,167,437 & 795,981 \\
Wikipedia & 1,000,951 & 758,228 \\
GlobalVoices & 147,829 & 116,360 \\
ELRC-CORDIS & 123,991 & 114,063 \\
NewsCommentary & 98,992 & 90,477 \\
\midrule
\textbf{Reference} & 555 & 498 \\
\bottomrule
\end{tabular}%
}
\caption{Data statistics showing the reduction from raw data to the final dataset used for retrieval. The cleaning pipeline included length filtering, deduplication, and linguistic filtering.}
\label{tab:corpus_stats}
\end{table}

\subsection{Extraction Method}

We extract domain-relevant sentences using embedding similarity with multiple centroids, following the intuition that crisis communication spans distinct registers that may occupy different regions of the embedding space.

\paragraph{Embedding and Clustering.} We encode both the reference and general corpus using paraphrase-multilingual-MiniLM-L12-v2, a multilingual sentence encoder \cite{reimers-2019-sentence-bert}. Since the reference corpus was designed to cover five main crisis scenarios, we cluster the 498 reference embeddings into k=5 clusters using k-means, generating five centroids that represent sub-domains within crisis communication.

\paragraph{Similarity Search.} For each sentence in the general corpus, we compute its cosine similarity to all five reference centroids and retain the maximum, allowing candidate segments to be matched to the most relevant crisis profile, and improving coverage of the resulting corpus. We use FAISS \cite{douze_faiss_2024} for efficient retrieval, extracting the top 50,000 most similar sentences from the general corpus, as candidates. 

\paragraph{Stratified Annotation.} To determine the optimal similarity cutoff, we ranked the candidate sentences by their maximum centroid similarity and divided the list into six partitions. We performed a stratified manual evaluation by randomly sampling and annotating 50 sentences from each partition as either in-domain or out-of-domain. The annotation was based on the hazard classification taxonomy established by the United Nations Office for Disaster Risk Reduction and the International Science Council \shortcite{undrr_isc_hips_2025}.

\paragraph{Threshold Selection.} Finally, we analyzed the domain relevance distribution across these partitions and established the final filtering threshold at the point where the proportion of out-of-domain sentences first exceeded that of in-domain examples. We retain some out-of-domain sentences in the upper ranks, as we posit that these borderline examples are beneficial. While they may not strictly describe a crisis event, both the human inspection and their proximity in the embedding space indicate that these sentences share register with the target domain. For this reason, we consider a soft domain boundary where these sentences are included and considered valuable examples for model training. As a result of the process, we retained 36,000 segments that we use for SFT training, as described in Section~\ref{sec:training}.

\section{Training}
\label{sec:training}
We perform Supervised Fine-tuning (SFT) on a \textbf{Tower-Plus-2B} model using parameter-efficient adaptation with QLoRA \cite{hu_lora_2021}. We intentionally opt for a compact language model rather than a larger LLM, as prior work has shown that for high-resource language pairs, domain adaptation yields diminishing returns beyond moderate model sizes when high-quality in-domain data is available \cite{pang_salute_2024,vieira_how_2024}. 

In crisis-response settings, smaller models are also better aligned with practical deployment constraints, including limited computational resources and low-latency requirements faced by public authorities and humanitarian organizations \cite{moorkens2025sociotechnical}. Moreover, compact models tend to exhibit more controlled generation behavior, reducing verbosity and the risk of hallucinations, which is particularly important in safety-critical communication where clarity and precision directly affect user comprehension and actionability.

\subsection{Paragraph Construction}

The model is fine-tuned on the crisis-domain corpus obtained through the retrieval pipeline described in Section~3. To better approximate the structure of authentic emergency communications, which typically consist of short informational blocks rather than isolated sentences, we adopt a paragraph-level training strategy.

Contextually similar sentence pairs are grouped into short paragraphs, generating approximately one paragraph every ten segments. Sentences within a paragraph vary in length and syntactic structure but are semantically coherent. The resulting training data therefore consists of a mixture of sentence-level and paragraph-level parallel examples, enabling the model to generalize across different granularity levels at inference time.

\subsection{Readability-Oriented DPO}
\label{subsec:dpo}
While translation accuracy is important, the actionability \cite{coche:hal-03295498} of crisis communications heavily depends on how comprehensible they are, when the target audience consists of non-native English speakers. To meet this requirement, we further optimize the model using Direct Preference Optimization \cite{rafailov_direct_2024} on a corpus of translations simplified to an English A2 proficiency level, as per the CEFR.

Preference pairs are constructed using synthetic simplified translations generated by \texttt{gpt-5-mini}, using a text simplification prompt used in relevant prior work \cite{barbu_easyjon_2025} and that we report in Appendix \ref{sec:simplification_prompt}. 

These simplifications aim to preserve semantic content while reducing lexical complexity and syntactic density, in line with established guidelines for emergency communication targeting linguistically diverse populations \cite{federici2019international}. To control for risks introduced by synthetic supervision, the simplified outputs undergo automatic quality checks using readability and adequacy metrics. In addition, a stratified sample is evaluated by human annotators to verify that simplification does not introduce semantic distortion or omit safety-critical information.

The resulting preference data biases the model toward translations that are both accurate and accessible, aligning optimization with the needs of L2 English readers in high-stakes emergency contexts.

\paragraph{Automatic Evaluation.}

\begin{table}[tbph]
\centering
\resizebox{\columnwidth}{!}{%
\begin{tabular}{lcc}
\toprule
\textbf{Metric} & \textbf{Baseline} & \textbf{Fine-tuned} \\
\midrule
Flesch Reading Ease $\uparrow$ & 30.21 & \textbf{45.04} \\
Flesch--Kincaid Grade $\downarrow$ & 15.56 & \textbf{12.65} \\
SMOG Index $\downarrow$ & 16.14 & \textbf{13.66} \\
Coleman--Liau Index $\downarrow$ & 13.32 & \textbf{11.33} \\
ARI $\downarrow$ & 17.12 & \textbf{13.94} \\
Dale--Chall $\downarrow$ & 11.31 & \textbf{10.13} \\
\bottomrule
\end{tabular}
}
\caption{Automatic evaluation on 1{,}000 test sentences, measured by readability metrics. Arrows indicate when higher or lower values are better.}
\label{tab:readability_dpo_results}
\end{table}

Table~\ref{tab:readability_dpo_results} reports automatic evaluation results on 1{,}000 test examples. The final model optimized with DPO model yields lower BLEU \cite{papineni_bleu_2002} and chrF \cite{popovic_chrf_2015} scores than the baseline, reflecting reduced surface similarity with the reference translations due to lexical simplification and paraphrasing. This is expected as the model was optimized to generate translations close to an English A2 proficiency level. While BLEU and chrF exhibit substantial drops, the actual trade-off remains comparatively small, as evidenced by COMET, which reports only a modest decrease (-6.8\%).

At the same time, the DPO model consistently improves all readability metrics, with a +14.8 increase in Flesch Reading Ease and substantial reductions in other indicators (Flesch-Kincaid, SMOG, ARI). These results indicate that the model optimization successfully reduces lexical and syntactic complexity, producing translations that are much easier to read for L2 English speakers.

\subsection{Ablation Study}
\label{sec:ablation}

To determine whether SFT is required to achieve both translation quality and readability improvements, we conduct an ablation study comparing DPO applied with and without prior domain adaptation.

We evaluate four configurations: (i) the base model without adaptation; (ii) the base model optimized directly with DPO; (iii) a model fine-tuned on the crisis-domain corpus using supervised learning only; and (iv) the full pipeline combining supervised fine-tuning and preference optimization.

Our findings, displayed in Table~\ref{tab:ablation_full} demonstrate that applying DPO directly to the base model yields large gains in readability but is accompanied by substantial drops across BLEU, chrF, and COMET, whereas applying DPO after supervised fine-tuning results in comparable readability improvements with small reductions in COMET.

\begin{table}[htbp]
\centering
\small
\resizebox{\columnwidth}{!}{%
\begin{tabular}{lcccc}
\toprule
\textbf{Training Configuration} &
\textbf{BLEU} &
\textbf{chrF} &
\textbf{COMET} &
\textbf{FRE} \\
\midrule
Base & 0.41 & 66.62 & 0.88 & 29.44 \\
SFT & 0.39 & 64.35 & 0.87 & 32.49 \\
Base + DPO & 0.07 & 31.10 & 0.73 & \textbf{74.40} \\
\midrule
\multicolumn{5}{l}{\textit{Optimal System}} \\
\midrule
\textbf{SFT + DPO} & \textbf{0.21} & \textbf{47.40} & \textbf{0.82} & \textbf{46.13} \\
\bottomrule
\end{tabular}
}
\caption{Overall performance and ablation study across four training configurations. 
The SFT+DPO model represents our final system. 
We report BLEU, chrF and COMET as quality metrics, and readability measured by Flesch Reading Ease (FRE).}
\label{tab:ablation_full}
\end{table}

\section{Evaluation}

To complement the quantitative results and better understand the practical implications of readability-oriented MT, we conduct a manual error analysis following the MQM framework \cite{lommel_tutorial_2018}, combined with Direct Assessment \cite{graham_accurate_2015}, using the platform Pearmut\footnote{https://github.com/zouharvi/pearmut}  \cite{zouhar2026pearmuthumanevaluationtranslation}. Two annotators, who were native speakers of the target language, proficient in the source language, and with expertise in translation studies,  applied the core set of MQM categories: accuracy, fluency, style, locale conventions, and verity, along with their respective subcategories. Errors were rated using four severity levels: trivial, minor, major, and critical, corresponding to weights of 0, 1, 5, and 25, respectively. In total, 250 segments were evaluated with the SFT model and the readability-oriented model respectively, resulting in 500 overall segments.

We note that our use of MQM follows a relatively strict error taxonomy, which leads to a high number of annotated errors, especially for outputs that deviate from more literal translation behavior. As such, MQM should be interpreted here as a fine-grained linguistic error analysis rather than as a direct proxy for overall user acceptability.

After annotating an initial set of 100 segments, inter-annotator agreement (IAA) was calculated to ensure the reliability of the annotations. The initial agreement, measured with Cohen's Kappa, was equal to $K=0.69$ in identifying the most common error category, which was accuracy. The mean absolute difference between the scores assigned by the two annotators was equal to 7.5. Most disagreements concerned the MQM categorization for simplification-related phenomena, particularly the distinction between Omission and Undertranslation. Following this preliminary phase, the annotators jointly reviewed the disputed cases, clarified the annotation guidelines, and reached consensus on the appropriate error labels. The remaining 400 translation segments were then annotated independently.

\newcommand{\bubblescale}{1.25} 

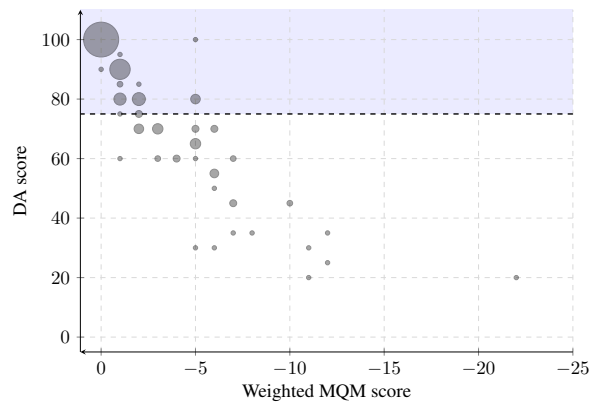
\begin{figure}[htbp]
\centering
\resizebox{\columnwidth}{!}{%
\begin{tikzpicture}
\begin{axis}[
    width=10cm,
    height=7cm,
    scale only axis,
    xlabel={Weighted MQM score},
    ylabel={DA score},
    xmin=-25, xmax=1.1,
    ymin=-5, ymax=110,
    x dir=reverse,
    xtick={0,-5,-10,-15,-20,-25},
    ytick={0,20,40,60,80,100},
    grid=major,
    grid style={dashed,gray!30},
    axis lines=left,
]

\begin{pgfonlayer}{background}
\path[fill=blue!8, draw=none]
    (axis cs:-25,75) rectangle (axis cs:1.1,110);
\end{pgfonlayer}
\addplot[dashed, thick, domain=-25:1.1] {75};

\pgfmathsetmacro{\sA}{\bubblescale*8.1}
\pgfmathsetmacro{\sB}{\bubblescale*4.7}
\pgfmathsetmacro{\sC}{\bubblescale*3.0}
\pgfmathsetmacro{\sD}{\bubblescale*2.8}
\pgfmathsetmacro{\sE}{\bubblescale*2.4}
\pgfmathsetmacro{\sF}{\bubblescale*2.2}
\pgfmathsetmacro{\sG}{\bubblescale*2.0}
\pgfmathsetmacro{\sH}{\bubblescale*1.6}
\pgfmathsetmacro{\sI}{\bubblescale*1.3}
\pgfmathsetmacro{\sJ}{\bubblescale*1.0}

\addplot[only marks,mark=*,mark size=\sA,draw=black,fill=black,fill opacity=0.35,draw opacity=0.45] coordinates {(0,100)};
\addplot[only marks,mark=*,mark size=\sB,draw=black,fill=black,fill opacity=0.35,draw opacity=0.45] coordinates {(-1,90)};
\addplot[only marks,mark=*,mark size=\sC,draw=black,fill=black,fill opacity=0.35,draw opacity=0.45] coordinates {(-2,80)};
\addplot[only marks,mark=*,mark size=\sD,draw=black,fill=black,fill opacity=0.35,draw opacity=0.45] coordinates {(-1,80)};
\addplot[only marks,mark=*,mark size=\sE,draw=black,fill=black,fill opacity=0.35,draw opacity=0.45] coordinates {(-3,70)};
\addplot[only marks,mark=*,mark size=\sE,draw=black,fill=black,fill opacity=0.35,draw opacity=0.45] coordinates {(-5,65)};
\addplot[only marks,mark=*,mark size=\sF,draw=black,fill=black,fill opacity=0.35,draw opacity=0.45] coordinates {(-2,70)};
\addplot[only marks,mark=*,mark size=\sF,draw=black,fill=black,fill opacity=0.35,draw opacity=0.45] coordinates {(-5,80)};
\addplot[only marks,mark=*,mark size=\sG,draw=black,fill=black,fill opacity=0.35,draw opacity=0.45] coordinates {(-6,55)};
\addplot[only marks,mark=*,mark size=\sH,draw=black,fill=black,fill opacity=0.35,draw opacity=0.45] coordinates {(-6,70)};
\addplot[only marks,mark=*,mark size=\sH,draw=black,fill=black,fill opacity=0.35,draw opacity=0.45] coordinates {(-7,45)};
\addplot[only marks,mark=*,mark size=\sH,draw=black,fill=black,fill opacity=0.35,draw opacity=0.45] coordinates {(-5,70)};
\addplot[only marks,mark=*,mark size=\sH,draw=black,fill=black,fill opacity=0.35,draw opacity=0.45] coordinates {(-2,75)};
\addplot[only marks,mark=*,mark size=\sH,draw=black,fill=black,fill opacity=0.35,draw opacity=0.45] coordinates {(-4,60)};
\addplot[only marks,mark=*,mark size=\sI,draw=black,fill=black,fill opacity=0.35,draw opacity=0.45] coordinates {(-3,60)};
\addplot[only marks,mark=*,mark size=\sI,draw=black,fill=black,fill opacity=0.35,draw opacity=0.45] coordinates {(-1,85)};
\addplot[only marks,mark=*,mark size=\sI,draw=black,fill=black,fill opacity=0.35,draw opacity=0.45] coordinates {(-7,60)};
\addplot[only marks,mark=*,mark size=\sI,draw=black,fill=black,fill opacity=0.35,draw opacity=0.45] coordinates {(-10,45)};
\addplot[only marks,mark=*,mark size=\sJ,draw=black,fill=black,fill opacity=0.35,draw opacity=0.45] coordinates {(-5,100)};
\addplot[only marks,mark=*,mark size=\sJ,draw=black,fill=black,fill opacity=0.35,draw opacity=0.45] coordinates {(-8,35)};
\addplot[only marks,mark=*,mark size=\sJ,draw=black,fill=black,fill opacity=0.35,draw opacity=0.45] coordinates {(-22,20)};
\addplot[only marks,mark=*,mark size=\sJ,draw=black,fill=black,fill opacity=0.35,draw opacity=0.45] coordinates {(-11,20)};
\addplot[only marks,mark=*,mark size=\sJ,draw=black,fill=black,fill opacity=0.35,draw opacity=0.45] coordinates {(-11,30)};
\addplot[only marks,mark=*,mark size=\sJ,draw=black,fill=black,fill opacity=0.35,draw opacity=0.45] coordinates {(-7,35)};
\addplot[only marks,mark=*,mark size=\sJ,draw=black,fill=black,fill opacity=0.35,draw opacity=0.45] coordinates {(-5,30)};
\addplot[only marks,mark=*,mark size=\sJ,draw=black,fill=black,fill opacity=0.35,draw opacity=0.45] coordinates {(-6,30)};
\addplot[only marks,mark=*,mark size=\sJ,draw=black,fill=black,fill opacity=0.35,draw opacity=0.45] coordinates {(-1,75)};
\addplot[only marks,mark=*,mark size=\sJ,draw=black,fill=black,fill opacity=0.35,draw opacity=0.45] coordinates {(-5,60)};
\addplot[only marks,mark=*,mark size=\sJ,draw=black,fill=black,fill opacity=0.35,draw opacity=0.45] coordinates {(-12,25)};
\addplot[only marks,mark=*,mark size=\sJ,draw=black,fill=black,fill opacity=0.35,draw opacity=0.45] coordinates {(0,90)};
\addplot[only marks,mark=*,mark size=\sJ,draw=black,fill=black,fill opacity=0.35,draw opacity=0.45] coordinates {(-2,85)};
\addplot[only marks,mark=*,mark size=\sJ,draw=black,fill=black,fill opacity=0.35,draw opacity=0.45] coordinates {(-1,60)};
\addplot[only marks,mark=*,mark size=\sJ,draw=black,fill=black,fill opacity=0.35,draw opacity=0.45] coordinates {(-1,95)};
\addplot[only marks,mark=*,mark size=\sJ,draw=black,fill=black,fill opacity=0.35,draw opacity=0.45] coordinates {(-12,35)};
\addplot[only marks,mark=*,mark size=\sJ,draw=black,fill=black,fill opacity=0.35,draw opacity=0.45] coordinates {(-6,50)};

\end{axis}
\end{tikzpicture}%
}
\caption{Relationship between weighted MQM score and DA score for the DPO model.
Bubble size reflects the frequency of the segments.
The shaded region (DA $\geq 75$) highlights translations judged high quality by DA despite MQM penalties.}
\label{fig:dpo_bubble_mqm_da}
\end{figure}

\paragraph{Results.} Our results find that the DPO model, optimized for readability, produces mostly accurate translations with a mean score of 83 points, compared to 95 for SFT. The translations produced by the DPO model were judged mostly acceptable, despite showing a higher number of minor and major errors compared to the SFT model. Particularly, 213 errors were annotated for the DPO model, and only 56 for the SFT one. However, 161 errors out of those 213 were considered minor in severity. This is substantiated by the relatively high DA scores assigned by the annotators.

Figure~\ref{fig:dpo_bubble_mqm_da} further illustrates the relationship between MQM and DA for the DPO model. Most segments are found in the upper-left region of the plot, indicating that many translations received high DA scores despite incurring minor MQM penalties. In particular, the shaded area (DA $\geq 70$) contains 113 segments, showing that a large proportion of the DPO outputs (56.5\%) were still judged high quality by human evaluators even when MQM identified minor issues. At the same time, the plot suggests a general correlation between the two measures: segments with heavily penalized MQM scores tend to receive lower DA scores. However, this tendency is not absolute, as several translations with moderately penalized MQM scores still achieve relatively high DA. This supports the view that MQM and DA capture overlapping but not identical aspects of translation quality. In particular, the minor MQM errors in these cases appear to be associated primarily with fluency, while DA traditionally evaluates how adequately the meaning the source sentence is expressed in the target \cite{bojar2017findings,graham_accurate_2015}. 

In the context of emergency communications, aimed at L2 English speakers, accuracy is arguably the most important quality dimension. The fact that many DPO translations remain highly rated in DA despite minor MQM errors therefore suggests that the model performs strongly for the communicative purpose of this task.

Furthermore, while SFT often outperforms DPO, over half of the evaluated segments received identical scores, indicating that DPO frequently produces translations of comparable quality despite its higher rate of errors, most of which are linked to simplification strategies. Although the gap in mean scores suggests systematic differences between the models, both systems consistently achieve scores in the upper half of the quality scale, indicating that overall translation quality remains high even when MQM annotations reveal a greater number of errors.

Consistently with the expected behavior of a model optimized for readability, most errors of the DPO model were found in the categories of Omission and Undertranslation, respectively with 43 and 74 annotated errors. For the SFT model, the most common error categories were Omission and Mistranslation. We include a detailed overview of the annotated errors for both models in Table 5 of the Appendix.

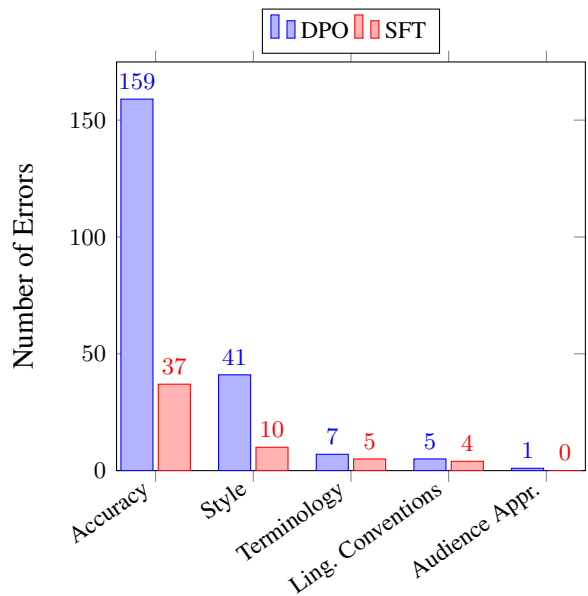
\begin{figure}
\centering
\begin{tikzpicture}
\begin{axis}[
    ybar,
    bar width=12pt,
    width=\linewidth,
    height=7cm,
    ylabel={Number of Errors},
    ymin=0,
    symbolic x coords={
        Accuracy,
        Style,
        Terminology,
        Ling. Conventions,
        Audience Appr.,
    },
    xtick=data,
    xticklabel style={
        rotate=35,
        anchor=east,
        font=\footnotesize
    },
    yticklabel style={font=\footnotesize},
    legend style={
        at={(0.5,1.02)},
        anchor=south,
        legend columns=2,
        font=\footnotesize
    },
    nodes near coords,
    nodes near coords style={font=\footnotesize},
]

\addplot coordinates {
    (Accuracy,159)
    (Style,41)
    (Terminology,7)
    (Ling. Conventions, 5)
    (Audience Appr.,1)
};

\addplot coordinates {
    (Accuracy,37)
    (Style,10)
    (Terminology,5)
    (Ling. Conventions, 4)
    (Audience Appr.,0)
};

\legend{DPO, SFT}

\end{axis}
\end{tikzpicture}
\caption{Distribution of dominant error categories per model. DPO shows substantially higher rates of errors related to its simplification behavior}
\end{figure}

\section{Conclusions}

In this study, we investigated the potential of using a hybrid data retrieval pipeline to collect relevant parallel data for domain adaptation, starting from a carefully curated small dataset. We used the retrieved data to fine-tune a small language model on the crisis domain, and then optimized via preferential learning to bias its outputs toward CEFR A2-level English.

Our results demonstrate that this two-stage optimization process enables translations that balance translation quality and accessibility, producing outputs that can be understood and acted upon by readers with diverse linguistic backgrounds. 

Automatic metrics show that the readability-oriented learning substantially improves textual accessibility across all readability metrics, while incurring only moderate decreases in COMET. The ablation study further confirmed that applying preferential learning without prior domain adaptation leads to severe degradation in translation quality, whereas combining SFT with DPO  yields improved readability with limited losses in semantic adequacy. Human evaluation confirmed our results, showing that our model can generate high-quality translations, while reducing unnecessary syntactic and semantic complexity.

In conclusion, our findings suggest that simplified English can function as a viable lingua franca for emergency communication in linguistically diverse contexts such as Italy. When fully multilingual dissemination is not feasible, domain-adapted MT systems optimized for readability can support the accessibility and actionability of emergency messages for speakers with varying levels of language proficiency.

\section{Future Work}
One of the main limitations of our study is the absence of a human evaluation explicitly targeting the acceptability of the generated translations. While we assessed translation quality and readability through automatic metrics and human evaluation, we did not directly measure whether the generated messages effective enable readers to act upon the information conveyed in the emergency warning messages.

Future work should therefore include qualitative human assessments involving L2 English speakers representative of the international communities in Italy. Such studies would assess not only perceived readability, but also clarity of instructions, accuracy in conveying key information, and the ability to identify recommended courses of action. Task-based evaluation, where participants are asked to interpret emergency messages and report their intended behaviors, could provide deeper insights into the communicative effectiveness of these translations.

In addition, future research should investigate whether simplification strategies can be further refined to minimize omission and undertranslation errors while preserving accessibility gains. Finally, while simplified English may function as a practical shared medium, investigating simplification across multiple languages could further enhance clarity in risk communication.

\section{CO2 Emission Related to Experiments}

Experiments were conducted using a private infrastructure, which has a carbon efficiency of 0.352 kgCO$_2$eq/kWh. A cumulative of 4 hours of computation was performed on hardware of type RTX 4090 (TDP of 300W). Total emissions are estimated to be 0.42 kgCO$_2$eq of which 0 percents were directly offset.

Estimations were conducted using the Machine Learning Impact calculator presented in Lacoste et al.~\shortcite{lacoste2019quantifying}.

\section*{Acknowledgments}
This work has been funded by the Italian National PhD programme in Artificial Intelligence, partnered by University of Pisa and University of Naples ``L'Orientale'', through a doctoral grant (ID 39-411-24-DOT23A27WJ-6603) established by Ex DM 318, of type 4.1, co-financed by the National Recovery and Resilience Plan. 

This work was also partially supported by the PhD Programme in Humanities and Technologies funded by the University of Macerata under D.R. No 253/2023.

The fourth author benefits from being member of the ADAPT SFI Research Centre at Dublin City University, funded by the Science Foundation Ireland under Grant Agreement No. 13/RC/2106\_P2.

\bibliography{custom,references}
\bibliographystyle{eamt26}

\appendix
\section{Appendix. Simplification Prompt}
\label{sec:simplification_prompt}

We report the prompt used in Section~\ref{subsec:dpo}, to ensure the replicability of the study. The prompt was used with the model gpt-5-mini with low reasoning effort, to generate the simplification corpus.

\begin{lstlisting}[style=promptstyle]
You are a text simplification AI.
Your task is to simplify the following input to A2 CEFR level. Use only common, everyday words that are appropriate for the context.
Choose words that native speakers would naturally use.
Explain essential terms if those can't be simplified and
maintain the content as in the original.

Input: {text}

Answer just with the simplification
and nothing else. Keep the original tone.
\end{lstlisting}

\section{Appendix. Annotated Examples}

Table~\ref{tab:mqm_appendix_combined} shows representative annotated translation examples drawn from the human evaluation dataset. The tables report outputs generated by the Supervised Fine-Tuned model and the readability-oriented DPO model.

For each segment, we provide the source text, the system translation, the identified MQM error category, the assigned severity level, and the final MQM score assigned by the annotators.

These examples illustrate the most recurrent error patterns discussed in Section~5, with particular emphasis on omission, undertranslation, and stylistic simplification phenomena.

Table~\ref{tab:mqm-errors} reports the distribution of MQM error categories, types, and severity levels for both the SFT and DPO systems.

\begin{table*}[t]
\centering
\scriptsize
\setlength{\tabcolsep}{3pt}

\begin{tabularx}{\textwidth}{
X
X
>{\raggedright\arraybackslash}p{2.8cm}
>{\centering\arraybackslash}p{1.2cm}
>{\centering\arraybackslash}p{0.8cm}
}
\toprule
\multicolumn{5}{c}{\textbf{SFT Model}} \\
\midrule
\textbf{Source (IT)} & \textbf{Output (EN)} & \textbf{Error Type} & \textbf{Severity} & \textbf{Score} \\
\midrule

Negli Stati Uniti, ad esempio, il 72\% delle donne con disabilità vive al di sotto della soglia di povertà. La discriminazione intensificata che le persone con disabilità possono subire a causa del loro sesso è particolarmente importante da considerare quando si tiene conto del fatto che, secondo l'Organizzazione per la cooperazione e lo sviluppo economico, le donne segnalano un'incidenza maggiore di disabilità rispetto agli uomini. & 
In the United States, for example, 72\% of women with disabilities live below the poverty line. The intensified discrimination that people with disabilities may face because of their gender is particularly important to consider when taking into account the fact that, according to the United Nations, women report a higher incidence of disabilities than men. &
Accuracy/Mistranslation 
&
Major &
65 \\
\addlinespace[1.5ex]

Negli ultimi anni abbiamo avuto modo di constatare il disastroso impatto che possono avere le alluvioni e anche il mio paese, che non per nulla si chiama “Paesi Bassi”, ha subito parecchie inondazioni. &
In recent years we have seen the disastrous impact that floods can have, and even my country, which is not called the Netherlands for nothing, has had several floods. &
Style/Language Resister 
Accuracy/Mistranslation
Accuracy/Mistranslation
&
Minor Minor Minor &
70 \\

\addlinespace[1.5ex]

Mettere sullo stesso piano "l'inquinamento di Internet", le emissioni di diossido di carbonio e la contaminazione dell'acqua aumenterebbe di gran lunga la possibilità di controllo delle operazioni online da parte del governo. &
Equating "internet pollution" with carbon emissions and water contamination would greatly increase the possibility of government control over online operations. &
Style/Language register  &
Minor &
90 \\
\addlinespace[1.5ex]

AC: Abbiamo previsto 8 priorità, che concernono ognuno di noi: la qualità dell'acqua, l'inquinamento della plastica, il trasporto marittimo, il turismo sostenibile, le piattaforme d'alto mare, le energie marine rinnovabili, il cambiamento climatico e la condivisione delle conoscenze sull'oceano: ocean literacy (alfabetizzazione sull'oceano). &
AC: We have identified eight priorities, which concern each of us: water quality, plastic pollution, shipping, sustainable tourism, offshore platforms, renewable marine energy, climate change and ocean literacy. &
Accuracy/Omission &
Major &
65 \\
\addlinespace[1.5ex]

In conclusione, credo sia necessario promuovere, e sostenere con un’organica attività di carattere formativo e con attrezzature e mezzi, l’azione del volontariato di protezione civile, che diventa una risorsa indispensabile per fronteggiare stati di emergenza e catastrofi naturali e di altri tipo. &
In conclusion, I believe that it is necessary to promote and support, through training and equipment, the action of civil protection volunteers, who are an indispensable resource for dealing with emergencies and natural and other disasters. &
Accuracy/Undertranslation &
Major &
80 \\

\midrule
\multicolumn{5}{c}{\textbf{DPO Model}} \\
\midrule
\textbf{Source (IT)} & \textbf{Output (EN)} & \textbf{Error Type} & \textbf{Severity} & \textbf{Score} \\
\midrule

\addlinespace[1.5ex]
Negli Stati Uniti, ad esempio, il 72\% delle donne con disabilità vive al di sotto della soglia di povertà. La discriminazione intensificata che le persone con disabilità possono subire a causa del loro sesso è particolarmente importante da considerare quando si tiene conto del fatto che, secondo l'Organizzazione per la cooperazione e lo sviluppo economico, le donne segnalano un'incidenza maggiore di disabilità rispetto agli uomini. & 
In the United States, for example, 72\% of women with disabilities live below the poverty line. The increased discrimination that people with disabilities may face because of their gender is especially important to consider when you consider that, according to the United Nations, women are more likely to have disabilities than men. &
Style/Awkward style
Accuracy/Mistranslation 
&
Minor Major &
55 \\
\addlinespace[1.5ex]

Negli ultimi anni abbiamo avuto modo di constatare il disastroso impatto che possono avere le alluvioni e anche il mio paese, che non per nulla si chiama “Paesi Bassi”, ha subito parecchie inondazioni. &
We have seen the impact of floods in recent years, and my country, which is called the Netherlands, has had a lot of flooding. &
Accuracy/Omission 
Style/Language register
Accuracy/Undertranslation 
Accuracy/Mistranslation
&
Minor Minor Minor Minor &
60 \\
\addlinespace[1.5ex]

Mettere sullo stesso piano "l'inquinamento di Internet", le emissioni di diossido di carbonio e la contaminazione dell'acqua aumenterebbe di gran lunga la possibilità di controllo delle operazioni online da parte del governo. &
Comparing Internet pollution to carbon emissions and water pollution would make it easier for the government to control the internet. &
Linguistic conventions/Punctuation 
Accuracy/Undertranslation 
Accuracy/Undertranslation    &
Neutral Minor Minor &
75 \\
\addlinespace[1.5ex]

AC: Abbiamo previsto 8 priorità, che concernono ognuno di noi: la qualità dell'acqua, l'inquinamento della plastica, il trasporto marittimo, il turismo sostenibile, le piattaforme d'alto mare, le energie marine rinnovabili, il cambiamento climatico e la condivisione delle conoscenze sull'oceano: ocean literacy (alfabetizzazione sull'oceano). &
AC: We have eight priorities, which are about each of us: water quality, plastic pollution, shipping, sustainable tourism, offshore platforms, renewable energy, climate change and ocean literacy. &
Style/Awkward style
Accuracy/Omission 
Accuracy/Omission
Accuracy/Undertranslation &
Minor Major Major Minor &
35 \\
\addlinespace[1.5ex]

In conclusione, credo sia necessario promuovere, e sostenere con un’organica attività di carattere formativo e con attrezzature e mezzi, l’azione del volontariato di protezione civile, che diventa una risorsa indispensabile per fronteggiare stati di emergenza e catastrofi naturali e di altri tipo. &
In conclusion, I believe that we must promote and support the work of civil protection volunteers, by providing them with the necessary training and equipment, so that they can be a valuable resource in the event of emergencies and natural or man-made disasters. &
Style/Language register
Accuracy/Omission
Accuracy/Addition &
Minor Minor Minor &
60 \\

\bottomrule
\end{tabularx}

\caption{Representative MQM error examples identified in the SFT and DPO model outputs.}
\label{tab:mqm_appendix_combined}
\end{table*}

\begin{table*}[ht]
\centering
\small

\begin{tabular}{llrrrrrrr}
\toprule
 & & \multicolumn{4}{c}{\textbf{DPO}} & \multicolumn{3}{c}{\textbf{SFT}} \\
\cmidrule(lr){3-6}\cmidrule(lr){7-9}
\textbf{Category} & \textbf{Type} & \textsc{maj} & \textsc{min} & \textsc{neu} & \textbf{Tot} & \textsc{maj} & \textsc{min} & \textbf{Tot} \\
\midrule
\multirow{5}{*}{\textbf{Accuracy}} & Addition & 3 & 2 & -- & 5 & 1 & 1 & 2 \\
 & Mistranslation & 15 & 18 & -- & 33 & 6 & 7 & 13 \\
 & Omission & 17 & 26 & -- & 43 & 4 & 5 & 9 \\
 & Overtranslation & -- & 4 & -- & 4 & -- & 2 & 2 \\
 & Undertranslation & 10 & 64 & -- & 74 & 3 & 8 & 11 \\
\cmidrule(lr){2-9}
\multirow{1}{*}{\textbf{Audience Appropriateness}} & Offensive & -- & 1 & -- & 1 & -- & -- & 0 \\
\cmidrule(lr){2-9}
\multirow{2}{*}{\textbf{Linguistic Conventions}} & Grammar & 1 & 3 & -- & 4 & -- & 4 & 4 \\
 & Punctuation & -- & -- & 1 & 1 & -- & -- & 0 \\
\cmidrule(lr){2-9}
\multirow{2}{*}{\textbf{Style}} & Awkward style & 2 & 14 & -- & 16 & -- & 4 & 4 \\
 & Language register & 1 & 24 & -- & 25 & -- & 6 & 6 \\
\cmidrule(lr){2-9}
\multirow{1}{*}{\textbf{Terminology}} & Wrong term & 2 & 5 & -- & 7 & 2 & 3 & 5 \\
\midrule
\multicolumn{2}{l}{\textbf{Total}} & 51 & 161 & 1 & 213 & 16 & 40 & 56 \\
\bottomrule
\end{tabular}
\caption{MQM error counts by category, type, and severity for DPO and SFT systems. \textsc{maj} = major, \textsc{min} = minor, \textsc{neu} = neutral. Note that SFT has no neutral errors.}
\label{tab:mqm-errors}
\end{table*}

\end{document}